\documentclass[letterpaper, 10 pt, conference]{ieeeconf}  

\IEEEoverridecommandlockouts                              

\usepackage{graphicx} 
\usepackage{wrapfig}
\usepackage{epsfig} 
\usepackage{mathptmx} 
\usepackage{times} 
\usepackage{amsmath} 
\usepackage{amssymb}  
\usepackage[ruled,linesnumbered]{algorithm2e}

{\end{list}}

\title{\Large \bf Adaptive Tensegrity Locomotion on Rough Terrain via Reinforcement Learning }

\author{David Surovik$^{1,*}$, Kun Wang$^{1,*}$, and Kostas E. Bekris$^{1}$
\thanks{*These authors contributed equally to this work}
\thanks{$^{1}$Department of Computer Science,
    Rutgers University, NJ, USA$\quad\quad$ 
    {\tt\small \{ds1417,kw423,kostas.bekris\}@cs.rutgers.edu}}
\thanks{Supported by NASA ECF grant \#NNX15AU47G to Kostas E. Bekris. The
authors would like to thank Jonathan Bruce and Massimo Vespignani from
the NASA Ames Intelligent Robotics Group for thoughts and information
regarding the SUPERball hardware.}
} 

\newcommand{\GPS}{\texttt{GPS}}
\newcommand{\ours}{\texttt{T6-GPS}}

\newcommand{\sa}{\mathbf r} 
\newcommand{\traj}{\tau}

\newcommand{\tgtdir}{\hat{\mathbf g}}

\begin{document}

\maketitle
\thispagestyle{empty}
\pagestyle{empty}

\begin{abstract}
The dynamical properties of tensegrity robots give them appealing
ruggedness and adaptability, but present major challenges with respect
to locomotion control.  Due to high-dimensionality and complex contact
responses, data-driven approaches are apt for producing viable
feedback policies.  Guided Policy Search (\GPS), a sample-efficient
and model-free hybrid framework for optimization and reinforcement
learning, has recently been used to produce periodic locomotion for a
spherical 6-bar tensegrity robot on flat or slightly varied surfaces.
This work provides an extension to non-periodic locomotion and
achieves rough terrain traversal, which requires more broadly varied,
adaptive, and non-periodic rover behavior.  The contribution alters
the control optimization step of \GPS, which locally fits and exploits
surrogate models of the dynamics, and employs the existing supervised
learning step. The proposed solution incorporates new processes to
ensure effective local modeling despite the disorganized nature of
sample data in rough terrain locomotion.  Demonstrations in simulation
reveal that the resulting controller sustains the highly adaptive
behavior necessary to reliably traverse rough terrain.

\end{abstract}


\section{Introduction}
\label{sec:introduction}
%
%
%

\begin{figure}[thpb]
  \centering
  \includegraphics[height=42mm]{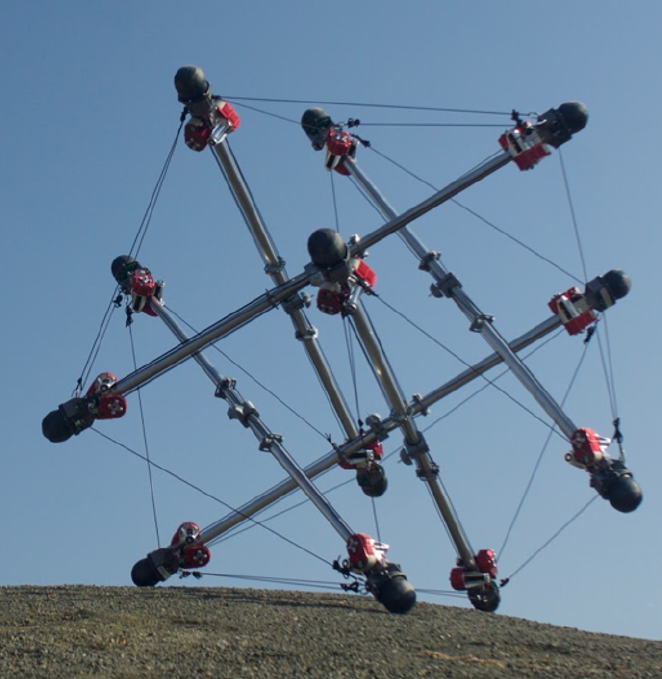}
  \includegraphics[height=42mm]{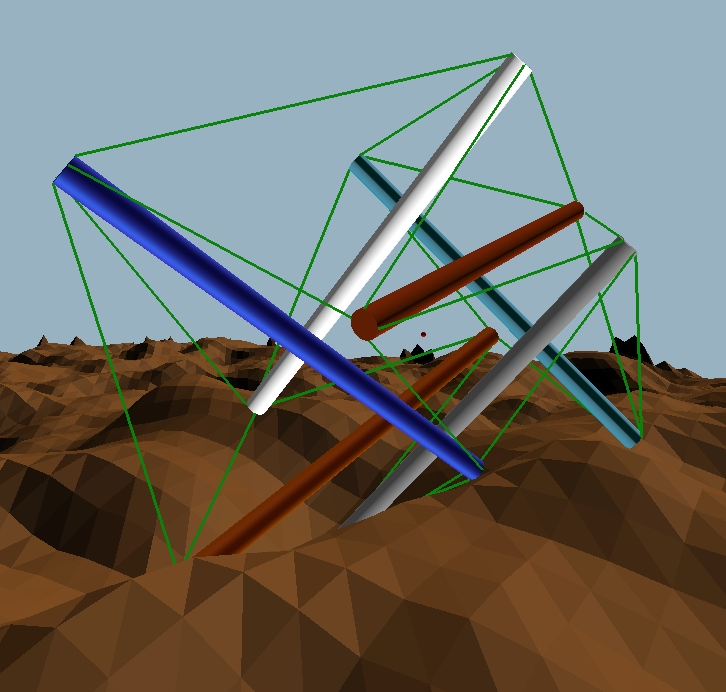}
  \vspace{-7mm}
  \caption{Left: hardware prototype of SUPERball version 2 at NASA Ames.  Right: Simulated behavior exiting a deep pit in rough terrain.}
  \vspace{-7mm}
  \label{fig:sbb}
\end{figure}

Tensegrity structures consist of rods suspended in a network of
elastic cables, so that all elements can freely pivot at the ``nodes''
where they are connected.  Force responses then occur as a compliant
reconfiguration of the structure, avoiding localized accumulation of
stresses and requiring less total material weight to withstand a given
load.  When granted the ability to change the lengths of their
elements, tensegrities can be made into robots that exhibit appealing
ruggedness and adaptability to rough terrain, such as NASA's 6-bar
tensegrity rover, SUPERball~\cite{v2hardware}, shown in
Fig.~\ref{fig:sbb}.  The same dynamical properties, however, also make
tensegrity locomotion control a hard and unintuitive
problem~\cite{MiratsTur2009, Rieffel2009}.

%


%

Deformation-based locomotion of tensegrities generally relies upon the
geometric relationship between the supporting base polygon and the
center-of-mass (CoM).  By deforming into a statically unstable
configuration, a ``flop'' forward onto an adjacent triangle can be
induced~\cite{Shibata2009}.  Hand-engineered methods may achieve this
by actuating just one or two elements, as has recently been
demonstrated in both software and hardware for ascending uniform
inclines as steep as $26^\circ$~\cite{chen_inclined_2017}.  Given a
model or database of the relationship between cable lengths and
vehicle shape, many-cable solutions can be discovered by search or
optimization~\cite{kimrobust, isrr_control, zhao2017efficient},
opening the door to more precise or adaptive behaviors.

\subsection{Related Work} \label{sec:relatedwork}

Dynamic tensegrity locomotion involves numerous complex factors, such
as a varying contact interface with the ground, that complicate the
use of model-based control.  Sampling-based motion planning, which has
been used for quasi-static deformation~\cite{Porta:2015aa}, can also
be applied to design kinodynamic paths that demonstrate the aptitude
of tensegrities on complex
terrain~\cite{littlefield_integrating_2016}.  The open-loop nature of
this approach, however, is inevitably fragile to errors, and it is too
computationally costly to be re-applied online.


This has motivated numerous investigations using data-driven control
methods, such as Central Pattern Generators
(CPGs)~\cite{MirletzSoftRobotics, Rennie:2018aa}, evolutionary
algorithms~\cite{iscen_learning_2015, Paul2006a}, and reinforcement
learning~\cite{zhang_deep_2017}.  CPGs can provide robust locomotive
gaits, but may limit the vehicle's ability to proactively adapt to the
environment with highly expressive shape changes.  Evolutionary
algorithms have been used to produce sustained locomotion on terrain
by a 6-bar tensegrity in simulation~\cite{iscen2014flop}, but
typically require very large datasets.

%

Reinforcement learning is often also associated with excessive data
requirements.  Guided Policy Search (\GPS) is a hybrid technique that
strongly reduces these requirements by exploiting gradient
information, but without the need for an apriori system
model~\cite{levine_learning_2014}. \GPS{} has been applied to produce
a periodic locomotive gait on the original hardware prototype of
SUPERball, which could actuate only half of its tensile
members~\cite{zhang_deep_2017}.  Similar behavior by a fully-actuated
6-bar tensegrity was also achieved in simulation on a surface with
height variations of a few percent of a bar length~\cite{agogino18}.



A commonality of prior controllers for 6-bar tensegrity
locomotion~\cite{chen_inclined_2017, iscen2014flop, zhang_deep_2017,
agogino18} is the regularity and periodicity of their behavior.  This
could limit the adaptiveness of the vehicle to increasingly
unstructured environments, where contact geometry becomes strongly
perturbed and the few discrete options for CoM motion may coincide
with obstacles.  For the two studies
using \GPS~\cite{zhang_deep_2017,agogino18}, this limitation is
related to the method's requirement for sample data to be organized
into local neighborhoods that correspond to linear time-varying
surrogate models of the dynamics.


This general limitation of \GPS{} has been identified and partially
addressed by ``Reset-Free'' \GPS{}, which utilizes clustering to
provide post-hoc localization of sample
data~\cite{montgomery_reset-free_2017}.  That modification, however,
targeted transient manipulation tasks and does not address the timing
variations involved in sustained non-periodic behaviors, such as
adaptive locomotion.  Combined with other limitations, this has
motivated recent work by the current authors in adapting the \GPS{}
pipeline to enable \emph{any-axis} locomotion of a 6-bar tensegrity on
a flat terrain, such that the CoM can follow arbitrary paths relative
to the contact geometry~\cite{anyaxis}.  The scope of that
experimentally-focused investigation, however, was restricted to
exploring the nature of any-axis motion on the plane, did not address
the case of rough terrain, and did not permit the detailed description or
evaluation of the algorithmic components that allow any-axis behavior
to emerge.


\subsection{Contributions and Outline}
%

%

This paper extends the line of work on adapting and employing the \GPS{}
reinforcement learning framework for any-axis planar locomotion with a
tensegrity robot~\cite{anyaxis}, as an example application domain that
involves complex dynamics and compliance.  Sec.~\ref{sec:background}
outlines \GPS{} and 6-bar tensegrity traits, along with a scheme for
symmetry exploitation that mitigates growth in sample complexity
relative to that of the more narrow-scope controllers in related prior
work~\cite{zhang_deep_2017, agogino18}.  The applied modifications in
the \GPS{} pipeline are described in Sec.~\ref{sec:approach} including
dimensionality reduction of the surrogate models, an alternate
post-hoc localization scheme relative to
``Reset-Free'' \GPS{}~\cite{montgomery_reset-free_2017}, and an
additional localization step introduced relative to prior work by the
authors~\cite{anyaxis}.

Experimental details are provided in Sec.~\ref{sec:setup}, including
description of a terrain environment of comparable or greater
difficulty than previous extremes~\cite{iscen2014flop,
chen_inclined_2017}.  Sec.~\ref{sec:results} demonstrates a controller
that successfully traverses this environment, while also exhibiting
any-axis characteristics.  The method will be referred to as \ours{},
to acknowledge its significant tailoring to the nature of 6-bar
tensegrities. Nevertheless, the discussion of
Sec.~\ref{sec:discussion} will address the more general lessons of
this investigation that can impact the deployment of reinforcement
learning pipelines, such as \GPS, to other highly complex and
dynamical systems.


%
%
%





\section{Background} 
\label{sec:background}


Let $\mathbf x' = \phi\left(\mathbf x,\mathbf u\right)$ describe the
discrete-time, nonlinear system dynamics as a function of the state
$x$ and controls $u$.  With observation $\mathbf y\left(\mathbf
x\right)$, define a control policy $\mathbf u
= \pi_\theta\left(\mathbf y\right)$ governed by a parameter vector
$\theta$.  The closed-loop dynamics are then $ \mathbf x'
= \phi\left(\mathbf x, \pi_\theta\left(\mathbf y\left(\mathbf
x\right) \right) \right) $.  Using the combined state/action vector
$\sa=\left[\begin{array}{cc} \mathbf x^T & \mathbf
u^T\end{array}\right]^T$, a trajectory
$\traj= \left[\begin{array}{cccc} \sa_0, \sa_1, \ldots, \sa_{T-1}\end{array}\right]$
is the length-$T$ sequence generated by $\phi$ and $\pi_\theta$ for an
initial state $\mathbf x_0$. Finally, let the running cost
$l\left(\sa\right)$ be the performance metric for any given time step
of controlled dynamics.

%
%

\subsection{Guided Policy Search}


The term \emph{Policy Search} refers to a class of algorithms that
calibrate a parameterized control policy by searching the space of
parameter values $\theta$.  Under a dataset consisting of $N$
trajectories of length $T$, this corresponds to the objective
\begin{align} 
    \theta^* = \texttt{argmin}\, \Sigma_{i=0}^{N-1}\Sigma_{t=0}^{T-1} l\left(\traj_i\left(t\right)\right).
\end{align}


As the complexity of the control policy architecture increases,
evolutionary algorithms and other black-box optimization approaches to
discover $\theta^*$ require significant amounts of sample
data. \emph{Guided Policy Search} (\GPS) is a technique designed to
train an artificial neural network representation of $\pi_\theta$ with
only moderate sample complexity and no requirement of an explicit,
differentiable dynamics model~\cite{levine_learning_2014}. The method
combines control optimization (the C-step) and supervised learning
(the S-step).

Key to the sample efficiency of \GPS{} is its exploitation of system
gradient information within the C-step.  In the absence of an apriori
model, gradients of $\phi$ and $\pi_\theta$ with respect to the
state-action $\sa$ are approximated by fitting linear time-varying
surrogate models $f$ and $p$ to the sample data:
\vspace{-.05in}
\begin{align}
    \mathbf x^\prime = f \left(t,\sa\right) &= F(t)\sa + \mathbf
    f(t) \\
    \mathbf u = p \left(t,\mathbf x\right) &= P(t)\mathbf x + \mathbf p(t)
\vspace{-.15in}
\end{align}
With the assumption that $l\left(\sa\right)$ is analytic, it is then
possible to compute an improved local policy $p^*(t,\mathbf x)$ via
the iterative Linear Quadratic Gaussian
(iLQG)~\cite{tassa_synthesis_2012} using differentiable $l(t,\sa)$ and
$f(t,\sa)$.  Applying this policy to the original state sequence
produces variationally improved actions $\mathbf u^*$.

\begin{wrapfigure}{R}{0.22\textwidth}
\vspace{-0.1in}
\centering
\includegraphics[width=0.21\textwidth]{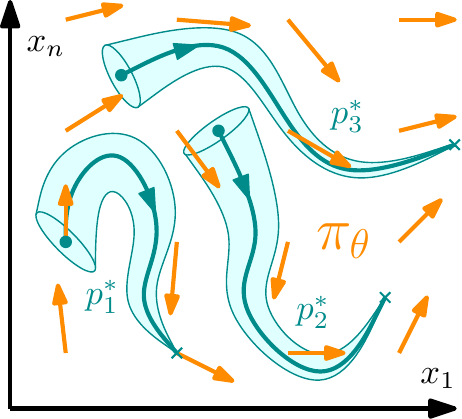}
\vspace{-0.1in}
\caption{The ``S-step'' of \GPS{} encodes multiple local controllers
$p_j^*$ into one global controller $\pi_\theta$.}
\label{fig:gps}
\vspace{-0.1in}
\end{wrapfigure}

Because both $\phi$ and $\pi$ are nonlinear, their surrogate models
have only limited local validity.  Several of these models $m_j
= \left(f_j,p_j\right)$ may then be necessary for improving feedback
actions for a large region of the state space.  The S-step consists of
using a globally accumulated set of observations and corresponding
locally improved actions, $D=\left\{\mathbf y, \mathbf u^*\right\}$,
for supervised training of the policy, i.e., learning $\theta$.
Fig.~\ref{fig:gps} illustrates this step as the matching of the global
policy field $\pi\left(\mathbf x\right)$ to a set of local policies
$p_j^*$ that funnel nearby states along a targeted path.


\vspace{-.15in}
\begin{algorithm}[h!]
\DontPrintSemicolon
\caption{\GPS{} Iteration}
\label{alg:gps}
$D \gets \emptyset$\;
\ForEach{$\mathbf x_0 \in \mathcal X_0$}
{
    $\mathcal T \gets$ RunSamples($\mathbf x_0, \theta_{i-1}, N$)\;
    $m \gets$ FitLocalModel($\mathcal T$)\;
    $\mathcal T^* \gets $ LQGBackwardPass($\mathcal T, m, l\left(\sa\right) $)\;
    $D \gets D\; \cup $ GetObservationActionPairs($\mathcal T^*$)\;
}
$\theta_{i} \gets$ SupervisedLearning($\theta_{i-1}, D$)\;
\end{algorithm}
\vspace{-.15in}


Algorithm~\ref{alg:gps} outlines the high-level procedure for one
iteration of \GPS.  For each $x_0$ of a set of initial conditions
$\mathcal X_0$, the previous policy $\theta_{i-1}$ is executed $N$
times to generate sample trajectories $\traj_k \in \mathcal T$.  The
time varying local model $m$ is then fit and used to conduct a
backward pass, updating each $\mathbf u$ to $\mathbf u^*$.  These are
paired with corresponding observations in the dataset
$D=\left\{\mathbf y,\mathbf u^*\right\}$ to supervise learning of
$\theta$.

Convergence is aided by augmenting the cost function with a
KL-divergence term, $l'\left(\mathbf r, p\right) = l\left(\mathbf
r\right) + cKL\left(p^*||p\right)$.  With weight $c$, this penalizes
the difference between the policy $p^*$, which improves the cost, and
the surrogate model $p$ of the policy that generated the original
trajectory.
%

\subsection{System Description} \label{sec:superball}

\begin{figure}[thpb]
  \centering
  \vspace{-2mm}
  \includegraphics[height=27mm,trim={0mm -15mm 0mm 0mm},clip=true]{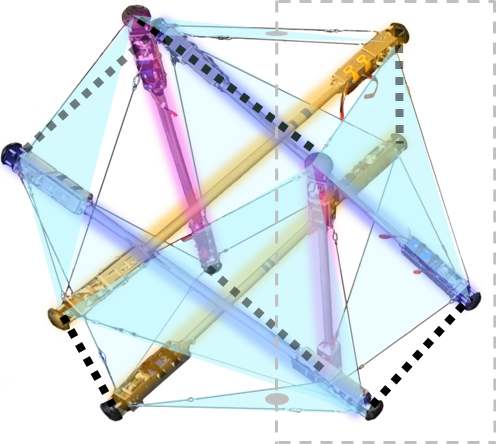}
  \includegraphics[height=27mm]{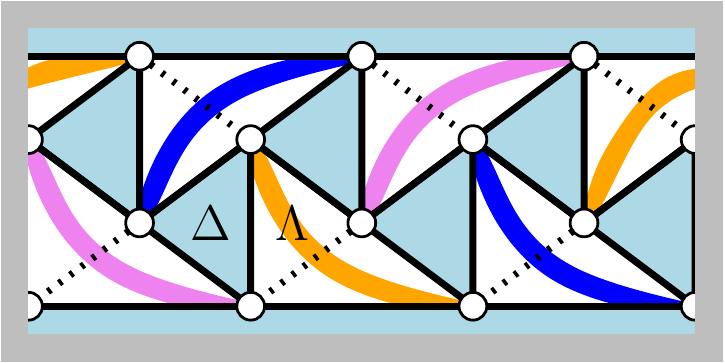}
  \vspace{-3mm}
  \caption{Topology of a 6-bar icosahedral tensegrity shown in 3D and 2D, using a cutting section shown in gray.  Solid lines indicate cables; dotted lines are virtual edges. $\Delta$ type triangles are shaded blue, $\Lambda$ unshaded.}
  \label{fig:topology}
  \vspace{-2mm}
\end{figure}

SUPERball is composed of 6 rigid bars, 1.94m long, that are isolated
in compression within a network of 24 actuated cables in tension.
This connectivity scheme is illustrated in Fig.~\ref{fig:topology},
which also highlights the 8 exterior triangular faces defined by three
cables ($\Delta$ type) While these triangles never share a common
edge, the remaining 12 triangles ($\Lambda$ type) occur in pairs that
share a ``virtual'' edge without a cable.

%

The state $\mathbf x$ consists of the 6-DoF rigid body state of each bar along with the rest length of each cable, which may differ from its actual length due to elastic deformation.
The control input $\mathbf u$ is the vector of desired cable rest lengths, while a separate control layer attempts to satisfy these specifications via motorized spools.
The simulation testbed representing $\phi\left(\sa\right)$ is the NASA Tensegrity Robotics Toolkit (NTRT)~\cite{SunSpiralSoftware}.

\begin{wrapfigure}{R}{0.20\textwidth}
\vspace{-0.1in}
\centering
\includegraphics[width=0.19\textwidth]{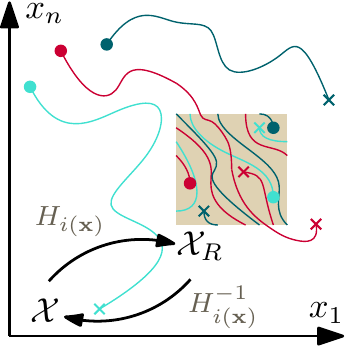}
\vspace{-0.1in}
\caption{Symmetry reduction of the state-action space volume.}
\label{fig:symmetry}
\vspace{-0.1in}
\end{wrapfigure}
In its neutral stance, SUPERball is a pseudo-icosahedron with 24th-order symmetry. 
Although deformation breaks spatial symmetry, each symmetric transformation can still be related to a permutation of the IDs of elements, which does not alter the vehicle's dynamics.
24 maps $H_i$ can then be defined that combine label permutation with a gravity-preserving orthogonal transformation, such that the intrinsic dynamics of the vehicle are not violated, i.e.,
$\phi\left(\mathbf r\right) = H_i^{-1}\phi\left(H_i\mathbf r\right)$.
%
Symmetry reduction is achieved by defining a rule $i(\mathbf x)$ such that all states are losslessly mapped into a volume 1/24th as large as the full space~\cite{asce_sym}.
The chosen rule produces one reference frame each for $\Delta$ and $\Lambda$ types of \emph{bottom triangles} (beneath the CoM) with fixed orientation.




\section{T6-GPS for Nonperiodic Locomotion}
\label{sec:approach}
\begin{figure}[thpb]
  \centering
  \vspace{-5mm}
  \includegraphics[width=0.48\textwidth]{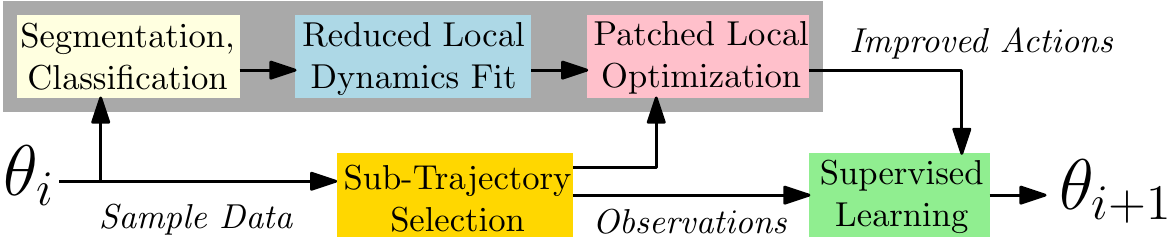}
  \vspace{-7mm}
  \caption{\ours{} iteration. Components within the gray box utilize dynamics reduction and post-hoc localization.}
  \vspace{-5mm}
  \label{fig:flowchart}
\end{figure}

The modified \GPS{} algorithm tailored for 6-bar tensegrities, \ours{}, 
is outlined in Fig.~\ref{fig:flowchart} 
and detailed in Algorithm.~\ref{alg:ours}.
\vspace{-2mm}
\begin{algorithm}[htb!] 
\DontPrintSemicolon
\caption{\ours{} Iteration}
\label{alg:ours}
$M, D_i \gets \emptyset$\;
$\mathcal T \gets$ RunSamples($\mathcal X_0, \theta_{i-1}$)\;
$S \gets $ SegmentSamples($\mathcal T$)\;
\ForEach{$\xi \in \Xi$}
{
    $S_\xi \gets $ FilterSegments($S,\xi$)\; 
    $M \gets M\; \cup $ FitLocalModels($S_\xi$)\;
}
\ForEach{$\tilde \tau \in$ \upshape{GetSubTrajectories}($\mathcal T$)}
{
    $\tilde m \gets $ PatchModelSeries($\tilde \tau, M$)\;
    $\tilde \tau^* \gets $ LQGBackwardPass($\tilde \tau, \tilde m, l\left(\sa\right) $)\;
    $D_i \gets D_i\; \cup $ GetObservationActionPairs($\tilde \tau^*$)\;
}
$\theta_{i} \gets$ SupervisedLearning($\theta_{i-1}, D_i$)\;
\end{algorithm}
\vspace{-5mm}


While Alg.~\ref{alg:gps} conducted repeated sampling of specific
initial conditions, which previous work~\cite{zhang_deep_2017,
agogino18} used to ensure periodic behavior, Alg.~\ref{alg:ours} does
not enforce any apriori structure on the sample set.
Instead, \emph{post-hoc localization} is conducted in lines 3-7, which
is important for accommodating any-axis and terrain-adaptive motion
that does not routinely repeat identical movements.  This process,
shown in the upper track of Fig.~\ref{fig:flowchart}, begins with
coarse localization through segment classification (lines 3-5,
detailed in Sec.~\ref{sec:segment}).  Further localization steps are
then taken in line 6, which corresponds to
Sec.~\ref{sec:warping}--\ref{sec:multimode}.  Lines 8-9 associate the
localized data back to original samples to allow completion of the
C-step, described in Sec.~\ref{sec:bwpass}.  Finally, lines 11-13
represent standard data accumulation and policy training.

%


%
\begin{figure*}[thpb!]
  \centering
  (a)\includegraphics[height=31mm]{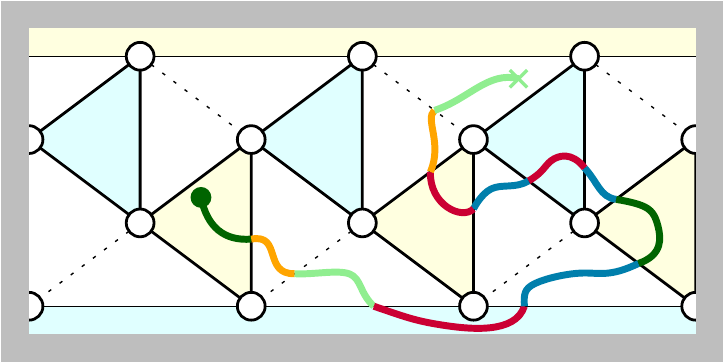}
  (b)\includegraphics[height=31mm]{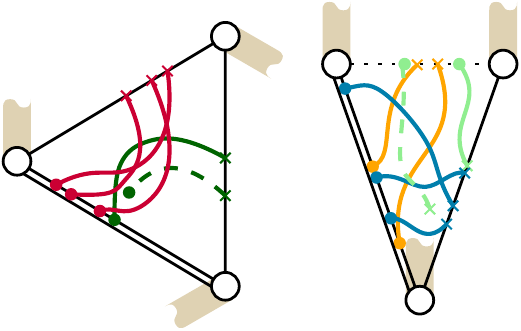}
  (c)\includegraphics[height=31mm]{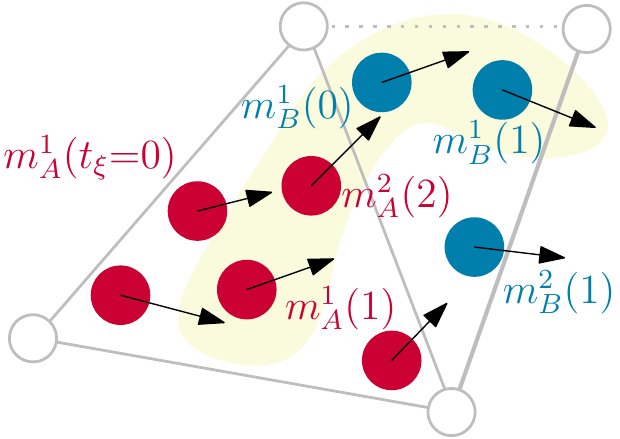}
  \vspace{-2mm}
  \caption{(a) segmentation of a single trajectory based upon CoM
  crossings over different bottom support triangles;  (b) classifying symmetry-reduced segments based on bottom triangle type and start/end edge relationships;  (c) fitting one or more surrogate models per time step per segment class. }
  \vspace{-4mm}
  \label{fig:segment}
\end{figure*}
\subsection{Segmentation and Classification} \label{sec:segment}

%
%

The first stage of post-hoc data localization breaks each trajectory
into several segments $ s = \tau\left(a \leq t < b\right)
= \left[\begin{array}{cccc} \sa_a, \sa_{a+1}, \ldots, \sa_{b-1}\end{array}\right] \label{eq:segment}
$, which are appropriate for a linear time-varying model.  A natural
criterion for this arises from the imposed scheme for symmetry
reduction into two bottom-triangle reference frames: segments are cut
when the bottom triangle ID changes, as illustrated in
Fig.~\ref{fig:segment}(a).  Segments are then classified via
Boolean-valued functions $\xi\left(s\right)$ such that FilterSegments
returns the segment set $\left\{s \in
S \;|\; \xi\left(s\right)\right\}$.  Fig.~\ref{fig:segment}(b)
illustrates segments within the symmetry-reduced space, colored based
on their class.


\subsection{Local Models - Time Warping} \label{sec:warping}
As is apparent in Fig.~\ref{fig:segment}(b), two segments may have similar characteristics while nonetheless occurring at different rates or durations.
%
Within FitLocalModels in Alg.~\ref{alg:ours}, \emph{time warping} is used to accumulate data points along a fixed-length time series $t_\xi=0,1,\ldots,T_\xi-1$.
Each $k$'th segment uses a time mapping $\tilde t_{\xi k}\in \mathbb N^{T_\xi}$ such that $s_k\left(\tilde t_{\xi k}\left(t_\xi\right)\right)$ returns the point from segment $k$ to be associated with the $t_\xi$-th time-step of the class.
Although more complex methods were considered~\cite{HACA}, it was found that a simple ``uniform stretch'' strategy functioned sufficiently, where for example a point halfway along a segment aligns to the time $T_\xi/2$.

\subsection{Local Models - Dynamics Reduction} \label{sec:dimreduce}

The symmetry-based reduction described in Sec.~\ref{sec:superball} shrinks the \emph{volume} of state space that must be covered by surrogate models and the global policy.
A further reduction, illustrated by the gray box in Fig.~\ref{fig:flowchart}, is applied to decrease the state \emph{dimension} accounted for in local fitting and optimization.
A linear map $L:\mathbf x\rightarrow \bar{ \mathbf x}$ is applied within FitLocalModels such that $m$ expresses $\bar{\mathbf x} = f\left(t,\bar{\mathbf r}\right)$ and $\mathbf u = p\left(t,\bar{\mathbf x}\right)$, where $\bar{\sa}=\left[\begin{array}{cc} \bar{\mathbf x}^T & \mathbf u^T\end{array}\right]^T$.
The purpose is to capture only the primary dynamical influences, so that the variational relationship between the control input and the resulting cost is coarsely but robustly approximated.  
Selection of $L$ must permit the evaluation $l\left(\bar{\mathbf x}\right)$, i.e., observability of the cost.

\subsection{Local Models - Multi-Modal Fitting} \label{sec:multimode}

The preceding steps produce one \emph{aligned} series of reduced
state-action sets for each class: $ \left[\left\{\bar{\mathbf
r}_0\right\}, \left\{\bar{\mathbf
r}_1\right\}, \ldots, \left\{\bar{\mathbf
r}_{T_\xi-1}\right\}\right]_\xi $. The next step is to produce
multiple linear models per step $t_\xi$.  The function recursively
applies RANSAC linear regression to produce models $m^h(t)$ such that
the $h$'th \emph{mode} is fit using the outliers of the $h{-}1$'th
mode.  Outliers are designated based upon a residual threshold equal
to one standard deviation of the dataset.  The hypothesis is that this
may account for different contact conditions between similarly shaped
states, while remaining less computationally intensive than fitting
Gaussian mixtures.


\subsection{Sub-Trajectory Backward Pass} \label{sec:bwpass}

%

Optimization requires dynamics gradients, provided from the set $M$ of
surrogate models, and cost gradients, obtainable analytically for each
trajectory.  The backward pass horizon is moderated by dividing each
trajectory into sub-trajectories $\tilde\tau$, which are essentially
longer segments, in GetSubTrajectories of Alg.~\ref{alg:ours}.  This
process keeps the horizon short enough to avoid excessive accumulation
of linearization error, but long enough to smooth out behavior across
transitions.

\begin{figure}[thpb]
  \vspace{-2mm}
  \centering
  \includegraphics[width=0.4\textwidth]{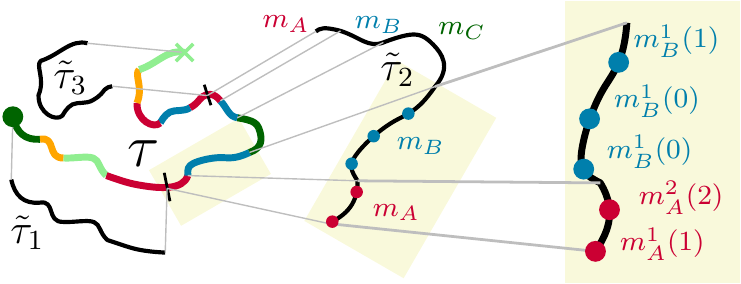}
  \vspace{-3mm}
  \caption{A single trajectory $\tau$ is broken into multiple subtrajectories $\tilde\tau$ (left).  Each point on a subtrajectory is associated with a surrogate model $m$ via segment classification (center), time warping and modality (right).}
  \vspace{-2mm}
  \label{fig:spans}
\end{figure}

Figure~\ref{fig:spans} illustrates the relationship between segments and sub-trajectories, along with the association of each point to a surrogate model $m^h_\xi(t)$.
Individual models are patched together by PatchModelSeries for use in
the BackwardPass, which returns $\tilde\tau^*$ with updated actions
$\mathbf u^*(t) = p^*\left(t,\mathbf x(t)\right)$.

%


%
%



\section{Experiment Setup}
\label{sec:setup}
\subsection{Scenario} \label{sec:scenario}

On each iteration of \ours, $N=300$ samples are executed for $T=250$
time steps each.  This corresponds to 25 seconds of robot behavior
under the applied 10Hz sampling frequency.  The initial state set
$\mathcal X_0$ was generated to randomize the vehicle's location,
orientation, and shape.


The rough terrain environment utilized is pictured in 3D in
Figs.~\ref{fig:sbb} and~\ref{fig:snapshots} and as a contour map in
Fig.~\ref{fig:groundtrack}.  The average slope of individual facets is
$38^\circ$ with standard deviation $16^\circ$.  Hills and troughs
exist on length scales similar to the size of SUPERball, with maximal
height variations of 1.55m, about 80\% of one bar length, which are
quite difficult to traverse.



The objective of each trial is to move the CoM through a sequence of
five randomly placed waypoints, which are half the diameter of the
vehicle.  If a waypoint is not reached within five seconds, the next
one is activated in order to prevent excessive time spent in stuck
states.  With $\tgtdir$ as the direction of the next waypoint, a
simple quadratic cost is defined in terms of center-of-mass velocity:

\vspace{-.2in}
\begin{align}
    l\left(\bar{\mathbf r}\right) &= \left(\mathbf v_{CoM}\left(\bar{\mathbf x}\right) - v^*\tgtdir\right)^TW\left(\mathbf v_{CoM}\left(\bar{\mathbf x}\right) - v^*\tgtdir\right) \label{eq:cost}
\end{align}
\vspace{-.2in}

\noindent where the weight matrix $W$ penalizes both ground-plane
components equally and de-weights the vertical.  A target velocity
$v^* = 0.8$m/s was chosen, which is faster than what can be achieved
uphill, but slower than tumbling downhill.


\subsection{Segment Classes}

%

%

%


Classification functions $\xi\left(s\right)$ evaluate the type of a
given trajectory segment according to multiple criteria:

\noindent 1) The category of the bottom triangle, i.e., $\Delta$ or
$\Lambda$, which represent fundamentally different configurations.

\noindent 2) The type of edge crossed upon entering a support triangle
and the type of edge crossed upon exiting it. For $\Delta$
triangles, the edge handedness influences the classification.  By
reasoning about the start/end edge relationship of a segment path,
relatively similar velocity values are achieved per class.  This
criterion also ensures the same type of final dynamics step, which
transfers the state to the next frame; see Fig.~\ref{fig:segment}(c).

\noindent 3) Distinguishing complete segments from ``partial''
segments, where either the beginning or ending of the segment is not
associated with an edge crossing. This occurs for the first and last
segments of each sample trajectory as well as for ``truncated''
segments, which represent the first portion of long segments where the
vehicle becomes stuck rather than reaching its next transition within
a reasonable time frame.  Classifying truncated segments allows the
algorithm to fit accurate surrogate models for improving actions that
previously failed to maintain motion of the platform.

Fig.~\ref{fig:segment}(b) visualizes some of the segment classes over
$\Delta$ and $\Lambda$ support triangles with different colors and
line types.

%
%

\subsection{Dynamics and Observation Spaces} \label{sec:setup_spaces}

Several definitions of the reduced state $\bar{\mathbf x}$ were
evaluated.  Each option included the $\mathbf v_{CoM}$ to ensure
compatibility with the gradient of the cost of Eq.~\ref{eq:cost}
during the backward pass.  Remaining contents were drawn from
different categories of full state variables: (1) the CoM-relative
positions of the lower six nodes, giving static stability and contact
information; (2) CoM-relative positions of all 12 nodes, providing
shape information; (3) the cable rest lengths, which express tension
if the shape is known; and (4) velocities of the nodes.

%

As outlined in Sec.~\ref{sec:superball}, the neural net input layer
accepts only symmetry-reduced sensor data.  It is assumed that the ID
of the bottom triangle can be classified from raw sensor data, as is
implemented on hardware~\cite{v2stepwise}.  The previous and current
ID are then used to determine the index of the symmetry reduction
transformation $H_i$ that expresses observations within the
appropriate reference frame.  The input layer is then provided with
(1) a Boolean indicating $\Delta$ or $\Lambda$ bottom triangle, (2)
the rest length for each of the 24 strings, (3) the 3D angular
velocity vector of each of the six bars, (4) the target direction
$\tgtdir$ expressed as an angle on the ground plane, and (5) the
``track'' ground plane angle giving the CoM velocity direction.  The
neural net itself is a simple fully-connected network with 3 hidden
layers and 256 neurons per layer.  Finally, the output cable lengths
$\mathbf u$ are relabeled for use with the actual state via $H_i^{-1}$
as in Fig.~\ref{fig:symmetry}.


\section{Results}
\label{sec:results}

\subsection{Flat Ground - Data Requirements vs. Standard \GPS}

Due to fundamental differences between periodic and nonperiodic
controllers, direct comparison of \ours{} to previous
approaches~\cite{zhang_deep_2017,agogino18} is not straightforward.
As a coarse comparison point, this work attempts to produce any-axis
CoM movement on flat ground without the use of post-hoc data
localization or symmetry reduction. This is to evaluate data
requirements of standard \GPS{} versus \ours.

Beginning at rest upon a particular $\Delta$ triangle with a fixed
orientation, the commanded direction of motion is discretized into
$J=36$ values, corresponding to the set of local models $m_j$.  For
each direction, $N=50$ samples of equal length $T=16$ are collected,
in keeping with the standard \GPS{} framework of
Algorithm~\ref{alg:gps}.  After $I=10$ \GPS{} iterations, the
controller was able to smoothly move the CoM in commanded directions
relative to the initial bottom triangle for one or two transitions ---
a result of limited scope for a tuned sampling cost of $IJNT=288,000$
total time steps.



This value already compares unfavorably to the $INT=125,000$ required
under an early version of \ours{} for this setup, an increase by a
factor of $20$ (the total number of triangles) would be necessary for
the standard \GPS{} pipeline to cover all transition cases.  Such
issues would only be compounded when training for rough terrain,
requiring another discretization dimension for $K$ types of terrain
features, with accompanying setup effort.  These findings motivate the
segment classification and time warping components of \ours{}
(\ref{sec:segment} and~\ref{sec:warping}), which naturally complement
the use of symmetry reduction, while more naturally and efficiently
covering the state space with long sample trajectories.

%
%

\subsection{Rough Terrain - Tuning and Cost Evaluation}

The remaining components of \ours{} are evaluated using the rough
terrain environment.  The backward pass horizon
(Sec.~\ref{sec:bwpass}) was not a sensitive parameter.  Sub-trajectory
lengths $10 \leq \tilde T \leq 20$ time steps (1 to 2 seconds), which
correspond to roughly one or two transitions, provided similar good
performance.  Results degraded for $\tilde T \leq 5$.


\begin{figure}[htpb]
  \centering
  \vspace{-4mm}
  \includegraphics[width=0.22\textwidth]{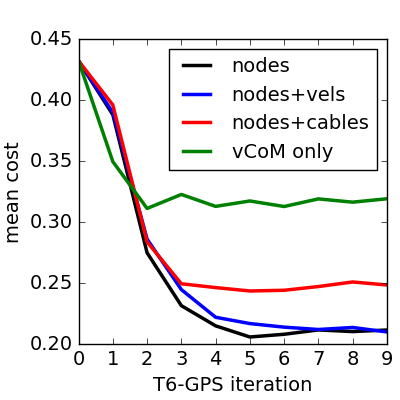}
  \includegraphics[width=0.22\textwidth]{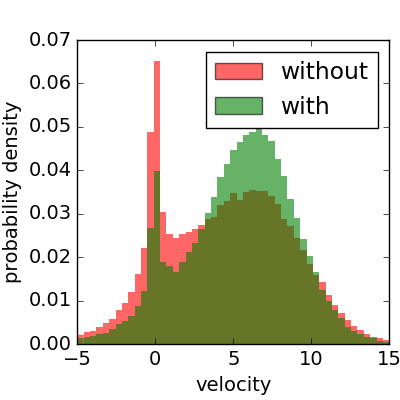}
  \vspace{-4mm}
  \caption{Left: performance of \ours{} for different surrogate spaces (mean of X runs). Right: velocity distributions without and with multi-modal fits.}
  \label{fig:costs}
  \vspace{-3mm}
\end{figure}

Modeling the complete state (node positions, velocities, and cable
rest lengths) resulted in memory issues due to excessive
dimensionality.  Figure~\ref{fig:costs} plots average sample cost
versus iteration count using smaller surrogate model spaces based upon
the value types listed in Sec.~\ref{sec:setup_spaces}.  When modeling
the cost state alone, $\mathbf v_{CoM}$, \ours{} could merely improve
the initial pure-noise policy to a motionless policy.  The inclusion
of node positions (corresponding to ``\emph{nodes}'' in the graph)
provided strong performance, converging to a value that will later be
shown to correspond to effective locomotion.  Additionally including
their velocities (``\emph{nodes+vels}'') marginally increased cost,
signaling diminishing returns in the trade off of additional
dimensionality incurred for additional information.  When instead
incorporating the rest lengths of the cables (``\emph{nodes+cables}'')
performance dropped significantly, suggesting that tensional effects
would require much narrower localization in order to be approximated
effectively.  Not shown, omitting the upper six node positions only
slightly increased the cost, highlighting the importance of interface
geometry over that of the general shape.

%
%
%



Noting that sample costs in Fig.~\ref{fig:costs} reflect the
performance of the previous iteration's controller, the iteration of
convergence is set to $i=5$ and the \emph{nodes} surrogate space is
chosen.  A total of $INT=375,000$ sampled time steps were used to
produce this controller, with a total runtime of about 1.0 hour on a
modern 8-core workstation.

To verify its benefit, multi-modal fitting (\ref{sec:multimode}) was
next disabled, resulting in a cost increase of more than 10\%.
Fig.~\ref{fig:costs}(b) plots the distribution of the forward speed
$\mathbf v_{CoM}\cdot\tgtdir$ with and without this feature, showing
that differing rates of stuck states (i.e., states with velocity close
to 0) are a major contributor to cost differences.  FitLocalModels
utilized 4 to 5 modes for the most common segment classes, and 1 or 2
for the least common.  Without multiple modes, the optimization step
may fail to account for contact differences between otherwise similar
states.

\begin{figure}[!thpb]
  \centering
  \vspace{-1mm}
  \includegraphics[width=0.48\textwidth,trim={0mm 1mm 0mm 13mm},clip=true]{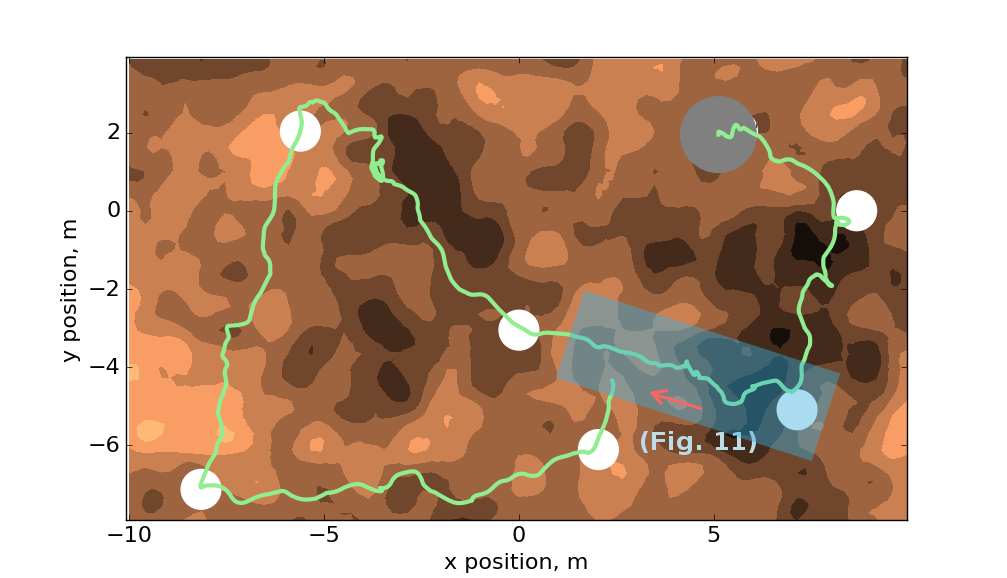}
  \vspace{-6mm}
  \caption{CoM ground-track originating at the gray circle (size of SUPERball) visiting several waypoints (white circles). Darker contours are lower terrain.}
  \vspace{-5mm}
  \label{fig:groundtrack}
\end{figure}

\subsection{Rough Terrain - Locomotive Behavior}

\begin{figure}[!thpb]
  \centering
  \vspace{-2mm}
  \includegraphics[width=0.48\textwidth,trim={0mm 0mm 0mm 10mm},clip=true]{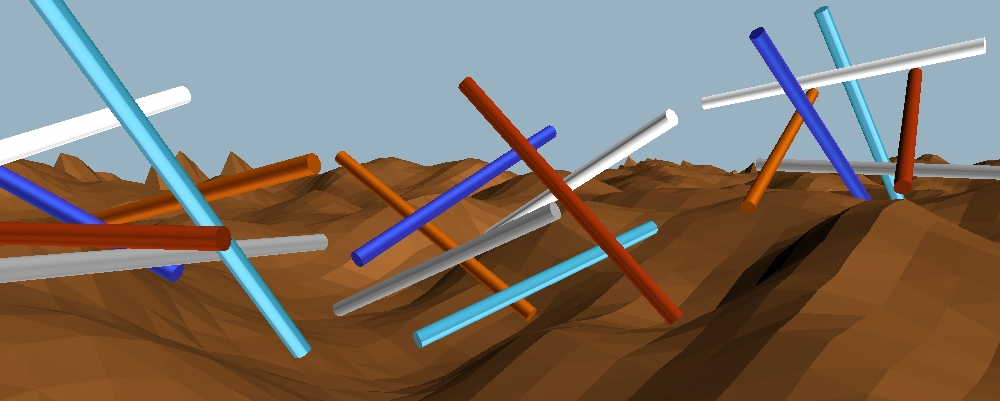}
  \vspace{-7mm}
  \caption{SUPERball traverses to the right across a difficult terrain feature.}
  \vspace{-2mm}
  \label{fig:snapshots}
\end{figure}

Behavior of the top-performing controller is now examined
qualitatively.  Figure~\ref{fig:groundtrack} gives a top-down view of
a typical CoM path lasting 1.5 minutes (900 steps).  The path is
generally smooth when motion is level or downhill, with some
irregularities or slower progress at uphill portions such as at
coordinates (5,-4) and (-3,1).  Figure~\ref{fig:snapshots} provides
snapshots of the (5,-4) hill ascent, which involves two steps up
locally steep bumps for a net height gain of more than 2/3 of a bar
length.  This feature thus has a similar average slope and overall
greater roughness relative to the previously attempted terrains
~\cite{iscen2014flop} (which maxed out at $18^\circ$).  Behavior is
far less constrained than in uniform slope
ascent~\cite{chen_inclined_2017}.

%

\begin{figure}[!thpb]
  \vspace{-2mm}
  \centering
  \includegraphics[width=0.48\textwidth]{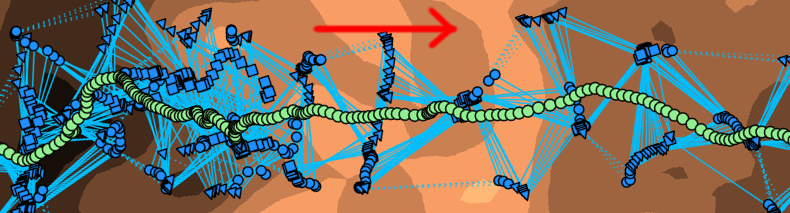}
  \vspace{-7mm}
  \caption{Detailed ground track during hill ascent. Geometry of the triangle below the CoM is shown in blue.}
  \vspace{-3.5mm}
  \label{fig:footprints}
\end{figure}

A zoomed-in view of the ground track in Fig.~\ref{fig:footprints}
provides details about the vehicle's geometry during this portion of
the trial.  At the right of the plot, where motion is relatively level
or downhill, the nonperiodic nature of the contact pattern reveals the
``any-axis'' characteristics of the controller.  This increased
freedom of directionality relative to most prior
work~\cite{iscen2014flop,zhang_deep_2017,agogino18} could potentially
be harnessed by a planner to navigate narrow feasible routes within an
especially hazardous landscape.

In the left of Fig~\ref{fig:footprints}, overlapping footprints and
broad movements of bottom nodes indicate significant slippage of
contacts. 
While the controller nonetheless resolves such features, this
friction behavior indicates some limits of the system itself, as also
noted in hardware and NTRT~\cite{chen_inclined_2017}. Alteration of
the friction coefficient within the NTRT environment did not resolve
the slippage, which may require compliant contact modeling for more
realistic behavior.  This would be in line with the ``softball''
attachments at the nodes of the latest SUPERball hardware
prototype~\cite{v2hardware}.

\section{Discussion and Future Work}
\label{sec:discussion}
%
%
%


Through the addition and modification of several components within the
\GPS{} framework, this work has demonstrated a sample-efficient way of
generating a feedback locomotion controller for a simulated 6-bar
tensegrity rover on rough terrain.  Results showed dynamic, adaptive,
and robust nonperiodic behavior for traversing highly nontrivial
features involving diverse contact geometries and configurations.


More broadly, this work illustrates that a combination of
dimensionality reduction and post-hoc data localization can greatly
improve the utility of surrogate models for optimizing sustained
adaptive behaviors on high-dimensional robots.  While the present
implementation of these steps is tailored to 6-bar tensegrities,
similar results might be achieved on other platforms via the same
principles.  For example, localization could be achieved with
automated segmentation and aligned cluster analysis, with variational
auto-encoding to reduce dimensionality.


By prioritizing the algorithmic discovery of adaptive feedback
behaviors, this work fits into the middle of a hierarchy of
considerations for useful deployment of mobile tensegrities.  The
level below is the transfer of such behaviors to hardware, which may
involve a reduction of sensor requirements and improved physics models
that capture the interaction of cables with convex terrain.  One level
above is integration with a planning algorithm, which addresses
controller limitations, such as occasional stuck states, but should
ideally have its role kept simple and lightweight by maximizing
controller utility and robustness.  These two objectives will steer
the direction of future work.

\clearpage

\bibliographystyle{IEEEtran}
\bibliography{icra19}

\end{document}